# Face editing with GAN's – A Review

Sarthak Mishra
*Dept. of Computer Science*
*MPSTME, NMIMS university*
Mumbai, India
sarthak.mishra25@nmims.edu.in

Parthak Mehta
*Dept. of Computer Science*
*MPSTME, NMIMS university*
Mumbai, India
parthak.mehta23@nmims.edu.in

Nikhil Chouhan
*Dept. of Computer Science*
*MPSTME, NMIMS university*
Mumbai, India
nikhil.chouhan63@nmims.edu.in

Neel Pethani
*Dept. of Computer Science*
*MPSTME, NMIMS university*
Mumbai, India
neel.pethani31@nmims.edu.in

Ishani Saha
*Dept. of Computer Science*
*MPSTME, NMIMS university*
Mumbai, India
ishani.saha@nmims.edu

***Abstract*— In recent years, Generative Adversarial Networks (GANs) have become a hot topic among researchers and engineers that work with deep learning. It has been a ground-breaking technique which can generate new pieces of content of data in a consistent way.**

**The topic of GANs has exploded in popularity due to its applicability in fields like image generation and synthesis, and music production and composition. GANs have two competing neural networks: a generator and a discriminator. The generator is used to produce new samples or pieces of content, while the discriminator is used to recognize whether the piece of content is real or generated. What makes it different from other generative models is its ability to learn unlabeled samples. In this review paper, we will discuss the evolution of GANs, several improvements proposed by the authors and a brief comparison between the different models.**

## I. Introduction

Machine learning has made new strides in its development with the help of deep learning. As a result, machines are now capable of performing tasks that not so long ago seemed impossible. We have seen computers beat the world champions of Go and chess, and are seeing impressive progress in speech recognition and image processing. Despite this progress, human ingenuity and creativity still hold the upper hand over that of machines. Developers have been working on finding solutions that do not sacrifice complexity or performance by instituting generative adversarial networks (GANs).

Generative Adversarial Networks (GANs) was proposed in 2014 by Goodfellow et al. [1] which introduced a novel two-player game framework that can be used to train generative models. The core idea of GANs is to utilize two competing neural networks, called generator (G) and discriminator (D) to compete against each other in the training process. The discriminator is composed of a convolutional neural network that is trained to distinguish between the real and the generated images. The generator is a neural network which is trained to generate data which maximizes a specific loss function. The generator takes random noise as input and tries to produce the most realistic content. The loss function is computed by the discriminator, and the generator is trained to minimize it. As the adversarial game goes on, the generator learns to produce data that the discriminator can't distinguish from real photos, and the discriminator learns to identify which images are real and which are fake. As they constantly compete, they eventually reach a stable point called a Nash equilibrium. A Nash equilibrium is when no competitor can improve its position by making any changes in strategy. When this point is reached, the GANs will stop generating new data. The generator learns a mapping between an input space and an output space of the model, while discriminator is trying to distinguish generated samples from real ones. The discriminator will provide unlabeled data to the generator. Under the theoretical framework of GANs, it theoretically guarantees that the generator will improve over time, and so far there is no other generative model that is guaranteed to improve like GANs.

Generative adversarial network (GAN) framework is based on the concept that generative model can be made better when exposed to errors or adversaries, because adversarial examples often can highlight deficiencies in the features extracted by a model. Originally, the GAN used the REINFORCE algorithm for training but over time many improvements have been made to it like neon-trigonometric functions(NTF) losses, negative log-likelihood loss, etc. The discriminator meets real data sample and synthetic data sample that were produced by the generator. Even though the generator does not meet real data sample, it learns by interacting with the discriminator. Both networks are trained to maximize probability of mistake by making less mistakes. As the discriminator learns to better categorize content, the generator evolves to better fool the discriminator. This evolution results in the creation of new and diverse pieces of content. The GAN architecture consists multiple layers - typically convolutional or connected layers.

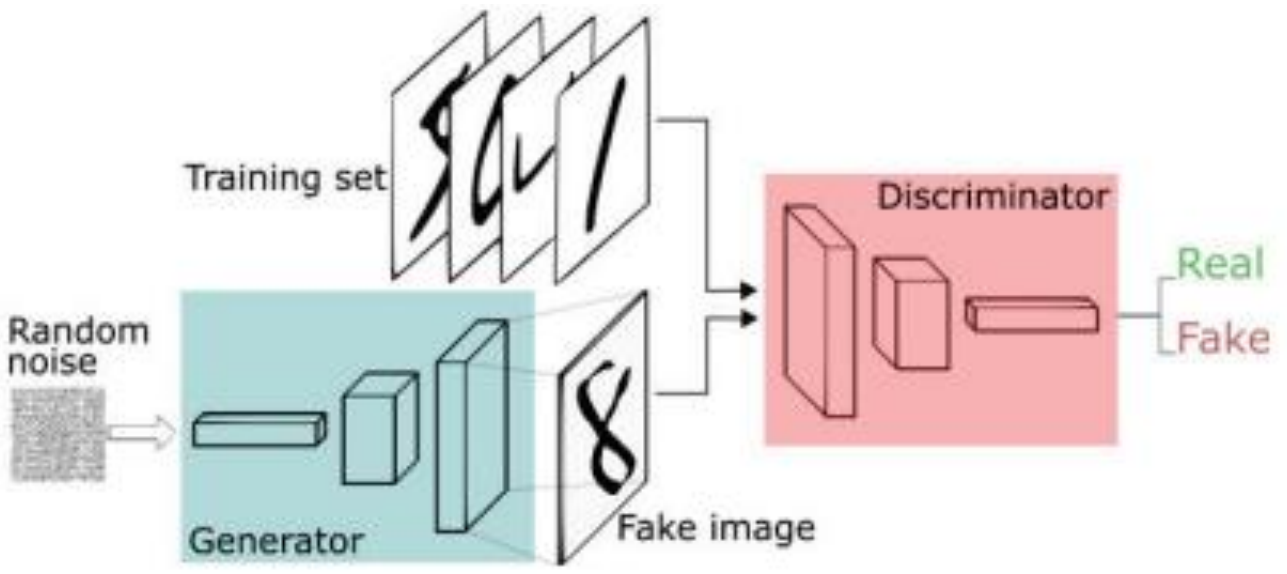

Figure 1: General Architecture of a GAN [10]

GANs are commonly used in unsupervised machine learning applications. The best-known application of GANs is in image generation, but extensions to video and audio are now being explored. GANs can learn to generate images conditioned on some observed data. In other words, you don't have to tell the generator what it should generate, nor do you have to give it an example of what you want it to generate. As a result, GANs can be useful in situations where labeled examples are scarce or expensive to obtain.

In this paper, we present a detailed discussion of Generative Adversarial Networks. Our focus is to provide a detailed discussion of what this notion entails and how it has been extended throughout history. The main parts in this article are listed as follows: in Section II, we investigated extensional models of GANs from the perspective of architecture. In Section III, we performed experiments on stylegan2 for image editing. In Section IV, we discuss the scores calculated during a short survey. We conclude in section V with an outlook into what is to come for generative networks.

## II. REVIEW OF LITERATURE

The increasing interest in generative adversarial networks (GANs) is driven by their potential to map latent variable spaces into data spaces and back through deep representations, and also by the potential ability to take advantage of large quantities of unlabeled image data. Interesting developments could arise from within the nuances of GAN training, and new opportunities for applications will likely grow with the power of deep learning.

In upcoming sections, we will discuss how GAN works and review the different methods based on generative adversarial networks (GANs) that are becoming available and some of the techniques which can scale up GAN with better training stability and increased image generation fidelity.

### *A. Generative Adversarial Nets*

Generative adversarial networks consist of 2 models that are placed in adversary to each other. The first is a generator, whilst the second is a discriminator. The generator tries to learn the data distribution and the discriminator tries to classify between fake images and real images, fake being from the generator and real being from the dataset. The loss is a minmax function in which the discriminator maximizes the loss which means that it correctly identifies the image as real or fake, and the generator minimizes the loss which means that the generator will fool the discriminator into identifying the fake images as real. It concludes with a comparison of GANs with Boltzmann machines and lists the advantages and disadvantages of the former. The Dataset used in this paper were MNIST, CIFAR10 and TFD. [1]

### *B. Generative Adversarial Networks: An Overview*

Without significant annotated training data, generative adversarial networks give a means to learn deep representation. They do so by backpropagating signals through a competitive process involving a pair of networks. These networks are known as generators and discriminators. Generators are represented by mapping to a space of data know as latent space, G:G(z)-->R|x| which generates random noise vectors. Discriminators are models which classify if the data is fake or real, D:D(x)-->(0,1). The discriminator discriminates the data generated by the generator from the ground trust and follows a feedback loop. Working in sync the combined model ends up as a GAN which generates real looking data. There are multiple architecture for a GAN such as Fully Connected GANs, Convolution GANs using CNN, conditional GANs, inference model GAN and Adversarial Autoencoders. This paper provides an overview of GANs and its architecture. It covers the training, formulation and the latent structure of a GAN. It also touches upon the applications of a GAN including image synthesis, semantic image editing, style transfer, image super-resolution and classification. The Dataset used in this paper were MNIST, CIFAR10 and TFD. [2]

### *C. Fader Networks: Manipulating Images by Sliding Attributes*

The encoder-decoder architecture presented in this paper is trained to reconstruct pictures by mapping image characteristics to the latent space. By changing the feature value in the latent space the image can be modified for particular application. The Fader network introduced in this paper is based on encoder-decoder architecture with domain-adversarial training on the latent space. The encoder is a CNN that maps an input image to its N-dimensional latent space. E:x-->R^n. The decoder is a deconvolutional network that produces a new version of the input data given its latent representation and data. D:(R^n,y)-->x. The loss function described is mean absolute or mean squared error to ensure the reconstruction matches the original data, future adversarial loss such as PatchGAN [4] can be used to obtain better texture and sharper images. The Dataset used in this paper were CelebA and Oxford-102. [5]

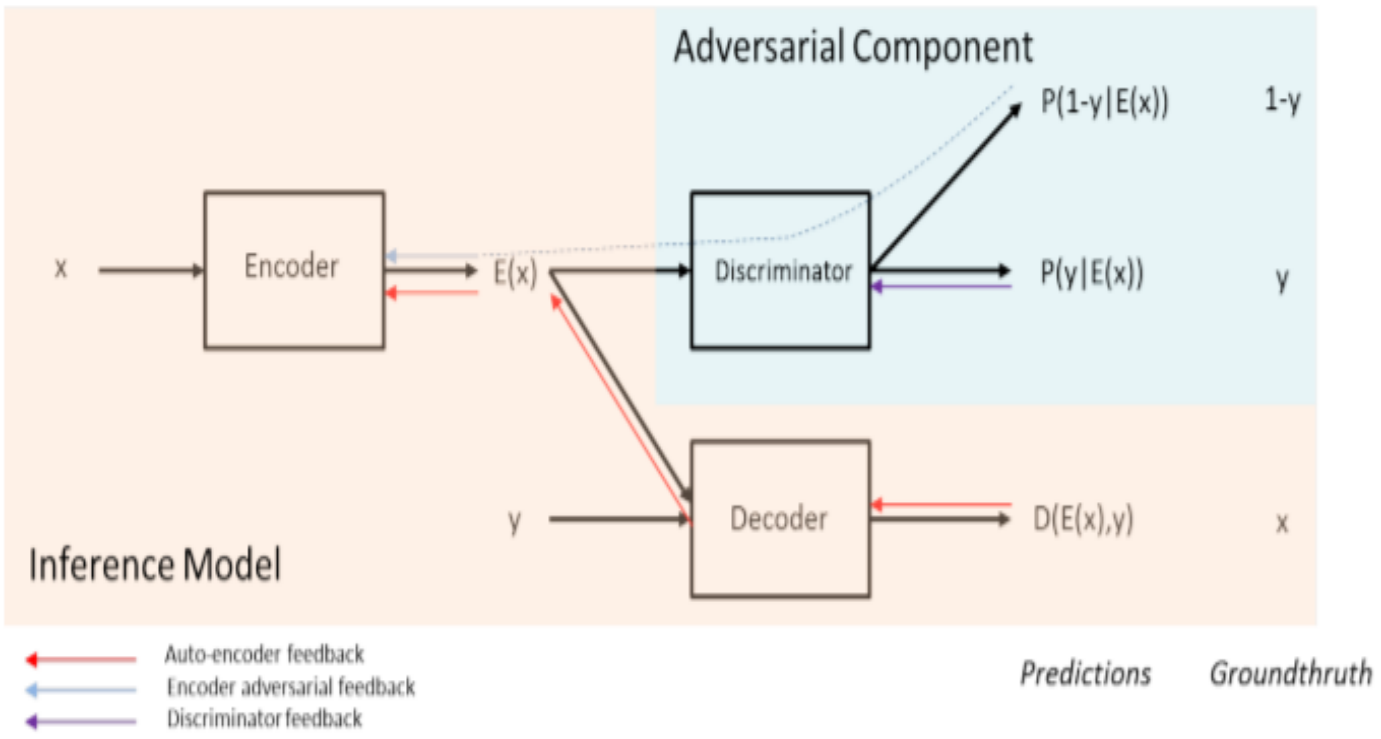

Figure 2: The basic architecture of Fader Network. [5]

*D. PuppetGAN: Cross-Domain Image Manipulation by Demonstration*

This paper proposes a model that can manipulate individual visual attributes of objects in real scene. The model learns to manipulate a specific property using synthetic demonstration, without any explicit labels. It can be applied to features such as shape, texture, lighting, other properties. The authors introduce a domain-agnostic encoder, a decoder for the real domain, and a decoder for the synthetic domain to disentangle the attributes which are to be controlled and all other present that are labeled or not present in the space. The model is trained on the dataset of sample data and area of interest data which is unlabeled of not present at all in the space. So, using the model input image can be manipulated with a synthetic reference image by applying a real decoder to attribute embedding of the reference image combined with the remaining embedding part of the input. The Dataset used in this paper were 300-VW, YaleB and MNIST. [3]

*E. A Style-Based Generator Architecture for Generative Adversarial Networks*

This paper introduces a completely new architecture for the generator in StyleGAN. In normal cases the generator starts from an input noise vector but in StyleGAN the generator starts from a constant learned matrix. Normal GANs use the latent vector directly but in StyleGAN the vector (Z) is first passed through a stack of fully connected layers which creates another vector (W). This new vector does not have to follow the same distribution as the input vector. For StyleGAN both vectors z and w have the same size-512. The vector w is then split into many styles by using learned affine transformations. These styles control adaptive instance normalization layers in the network after convolution layers. Along with these styles gaussian noise images are also fed to each layer. This approach disentangles many features and also improves the quality of the images. This model produces state of the art images which can be fined tuned later to adjust individual features without entanglement. The Dataset used in this paper was FFHQ. [6]

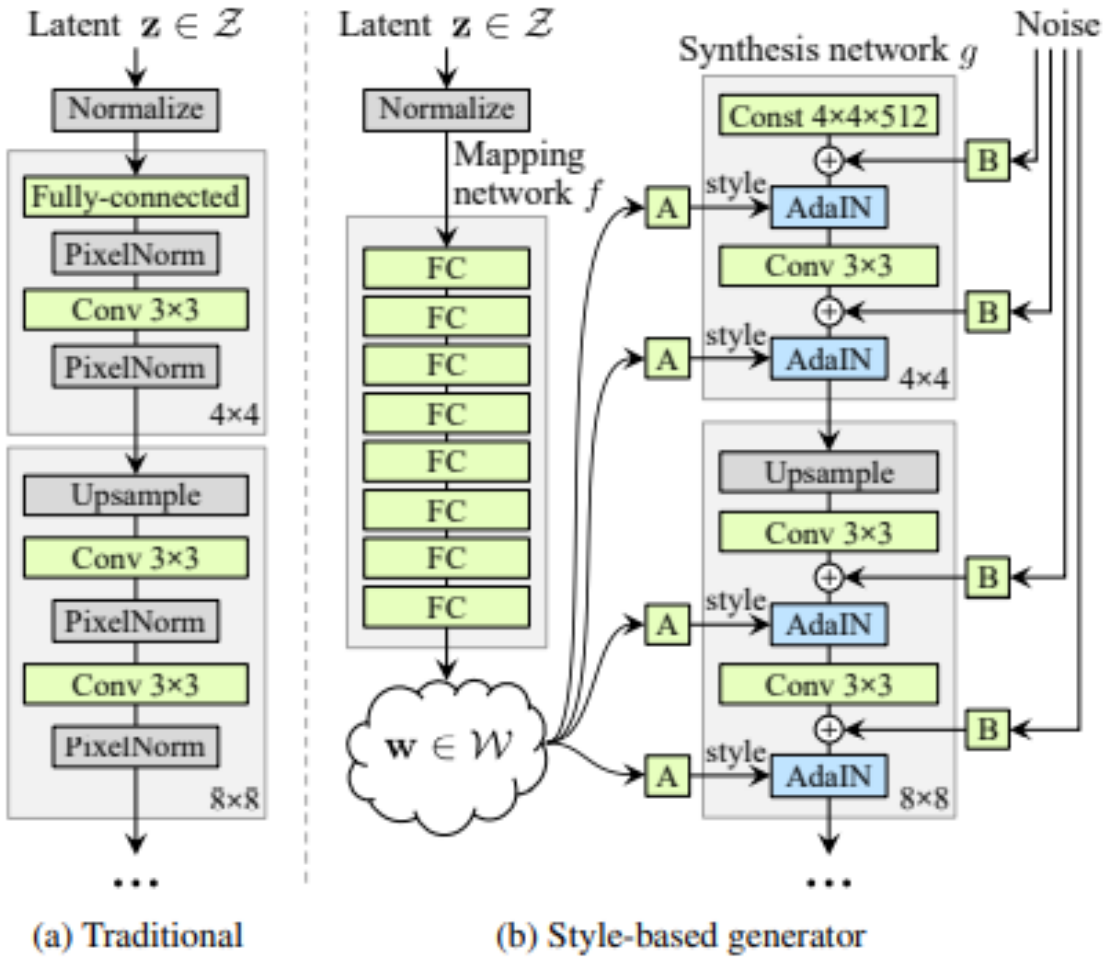

Figure 3: The basic architecture of StyleGan. Its generator has a total of 26.2M trainable parameters, compared to 23.1M in the traditional generator. [6]

*F. AgeGAN++: Face Aging and Rejuvenation with Dual Conditional GANs*

Face aging and rejuvenation, is a method for predicating what a person looks like in his different age stages. While previous work brought about a significant progress in the domain, there are still some problems that exist, for example, most prior works require sequential data during training, while it is very rare in existing datasets and rendering an aging face without neglecting the personality at the same time. To deal with these problems, the authors of this paper propose a novel dual conditional GANs mechanism, called AgeGAN, which enables face aging and rejuvenation to be trained utilizing multiple sets of unlabeled face images with different ages. The AgeGAN firstly converts a face image into other age stages based on a target age condition, then it learns to invert the task with the source age condition. However, the training process of the AgeGAN is complicated and the photo-realistic of the generated face images is required to be improved. Therefore, they propose an updated version of AgeGAN, termed AgeGAN++, which shares the weights between the primal part and the dual part to streamline the model. The Dataset used in this paper were UTKFace, FG-Net, CACD2000 and Morph. [7]

*G. Disentangled and Controllable Face Image Generation via 3D Imitative-Contrastive Learning*

In this paper, the authors investigate synthesizing face images with multiple disentangled latent spaces characterizing different properties of a face image. They then propose DiscoFaceGAN that generates realistic face images of virtual people with independent latent variables of identity, expression, pose, and illumination. The latent space is interpretable and highly disentangled, which allows precise control of the targeted images, that is degree of each pose angle, expression, lighting intensity and direction, achieving flexible and high-quality face image generation that is not achieved by any previous method. This method can also be used to embed real images into the latent space and edit the factors in a disentangled manner. The Dataset used in this paper were LFW and FFHQ. [8]

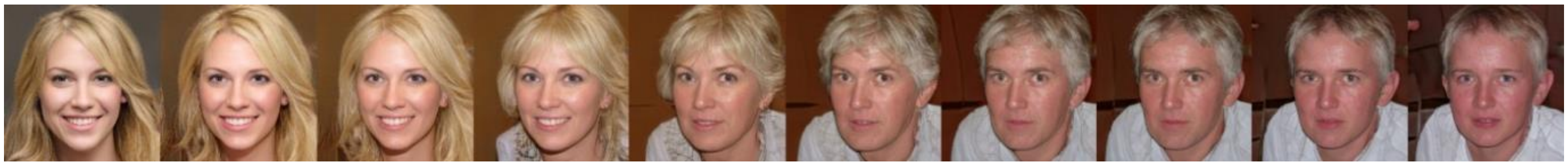

Figure 5: This figure show the interpolation between the 1st and last image. As we move from the left to the right, we can see that the image slowly starts to obtain the features of the last image and finally completely turns into the last image. The images in the center should contain features from both the images equally.

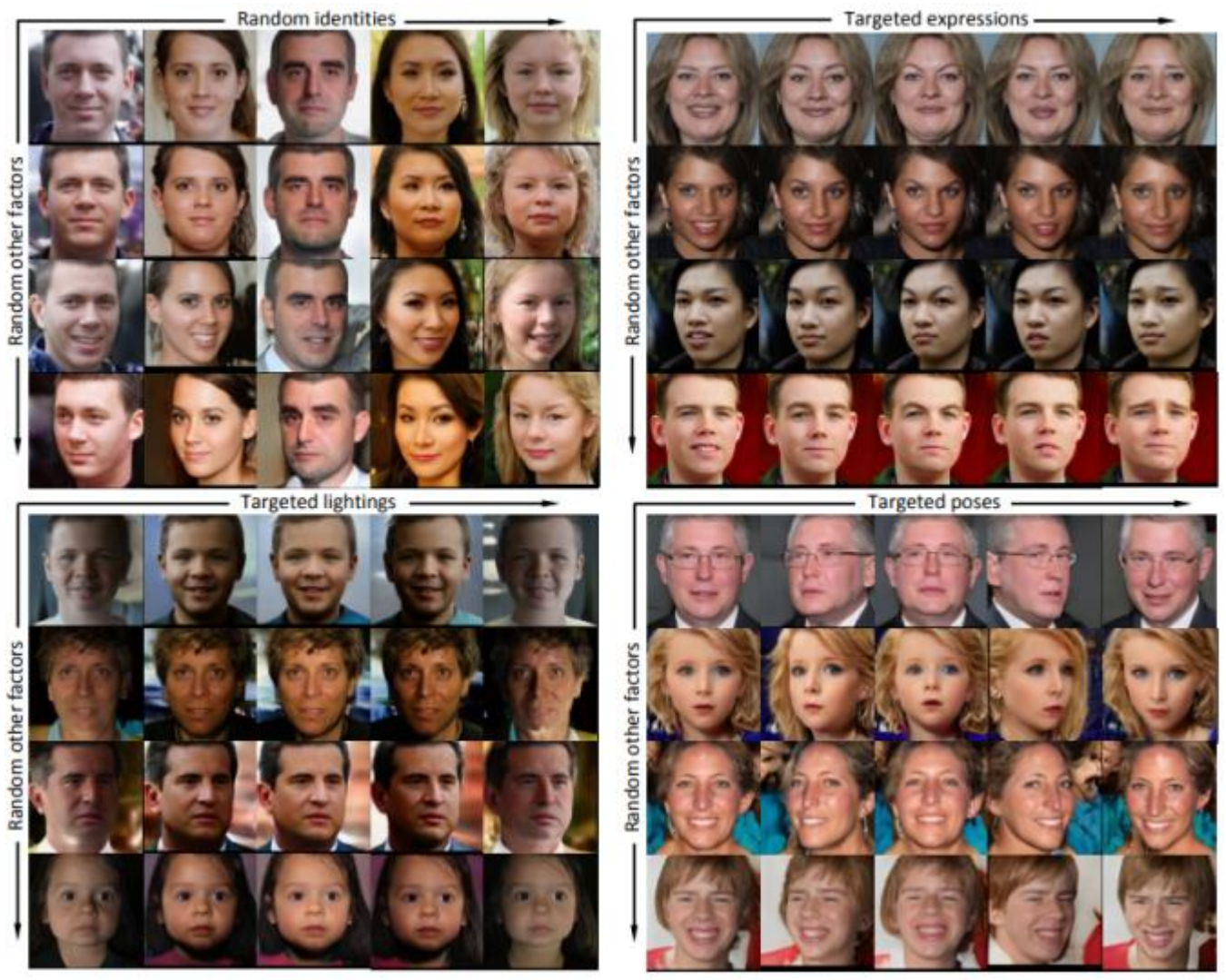

Figure 4: Face images generated by the DiscoFaceGAN. As shown in the figures, the variations of identity, expression, pose and illumination are highly disentangled, and we can precisely control expression, illumination and pose. [8]

### *H. Interpreting the Latent Space of GANs for Semantic Face Editing*

This paper addresses the latent space exploration issue that other GANs have. The authors have created a model by which latent space of a pretrained model can be explored and then each image can be edited for individual disentangled features. The authors have designed this model to explore how single or multiple semantics are learned by the GAN model and how they are represented in the latent space. The authors say that the GANs latent space creates subspaces in which these attributes are disentangled, and those subspaces can be used to interpolate between features. This GAN can be used with other pretrained architecture GANs. [9]

## III. Experiments on stylegan2

Stylegan2 is one of the current state of the art generative model released by Nvidia. As mentioned above it proposes a new architecture that leaves all the other the other models behind in terms of quality. Hence, we will be using it for our experiments. The experiments and their results have been explained below. GAN's generally map the dataset to a multidimensional distribution. To generate data, we can sample a random vector from this distribution and use that as input. The output image can be changed either completely of minimally by varying this, vector accordingly. Therefore, we can say that it is the vector that controls the attributes that we can find in the generated images. Randomly changing the vector is not useful, therefore techniques have to be developed to help change this vector in such a way that only a desired feature in the image changes and nothing else. For our experiments the models used were trained on FFHQ dataset.

### *A. Interpolation of latent vectors and Style-Mixing*

By passing a random vector through the StyleGan network, an image with certain features is generated. A latent vector is completely responsible for the kind of features the image will have. Each of these vectors can be considered as a point in the latent space of the GAN. In math, we can draw a line between 2 points and calculate all the points that lie on that line provided the coordinates of the 2 points and step size. Similarly, if we virtually plot multiple points in the latent space of the GAN, we should be able to connect these via lines and at the same time calculate all points that lie on those lines. All these new points are latent vectors and can generate images with certain features. For our experiment we took 2 vectors and calculated the line between them and also interpolated the points in between the two. The interesting part is more visible when all these 2 images are displayed side by side.

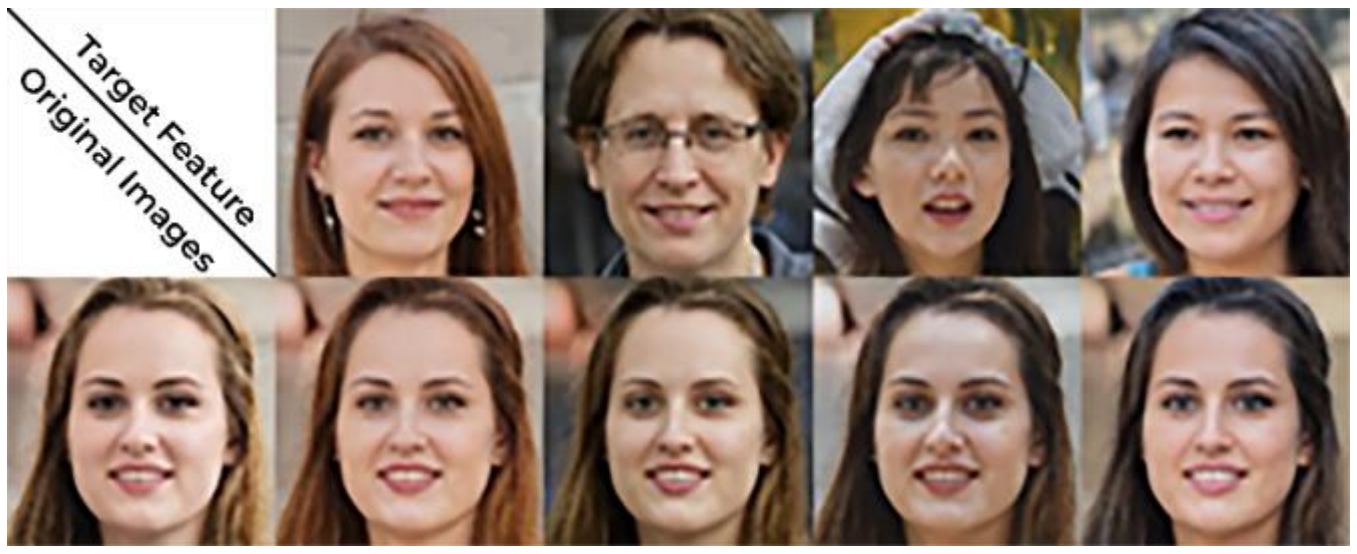

Figure 6: The above image showcases the style-mixing capability of stylegan 2. This image consists of 1 main image and 4 images from which features are extracted. In the above example, we can see that the result images contain major features like face shape, eyes expression etc. from the main image but contain small features like color, hair color etc. from the other 4 images.

One of StyleGan's feature is its mapping network. This network consists of 8 fully connected layers and converts an input vector z to a vector w. W can also be called as the style vector. The size of z and w vary with image size but for our experiments z was {1,512} and w was {18,512}. Style mixing is a method in which 2 w vectors are merged to generate an image consisting of the combination of the styles in the 2 images. This combination can be controlled by varying the amount of mixture of the 2 w vectors. For example, if w1 is a vector of a

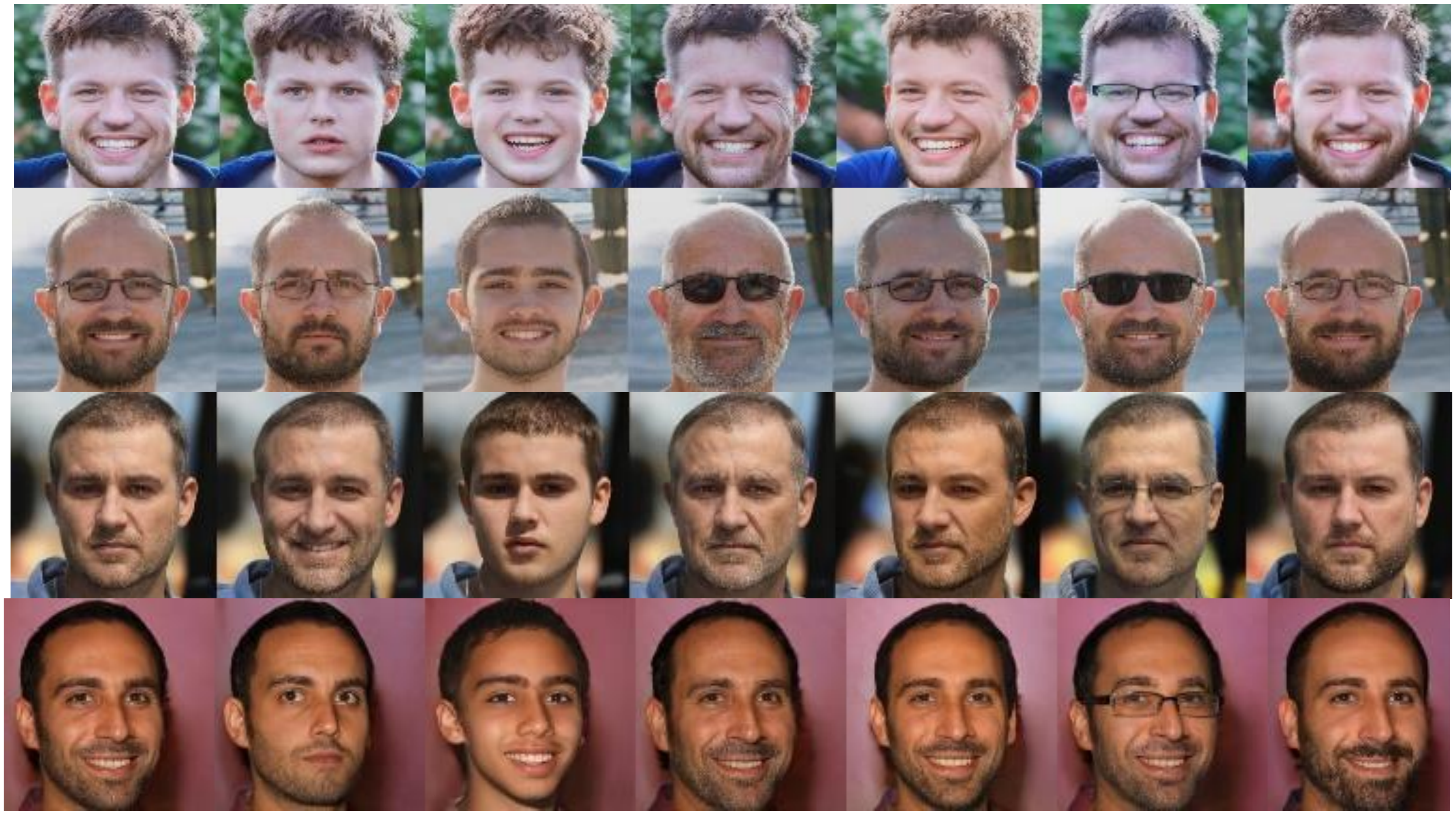

Original Smile Age(Young) Age(Old) Pose Glasses Beard

Figure 7: The above image shows the output of the feature directions being applied to sample images. As we can see this method works really well for some features and images. It is also successful in retaining most of the features from the original image.

male with brown hair and w2 is a vector of a female with black hair. If merged, the resulting image may be either of a male with black hair or a female with brown hair. This is controlled by carefully selecting the no of vectors from both w1 and w2, the position of the selected vectors and how to merge them. For example, we can take 9 vectors from w1 and 9 from w2. Now while selecting the vector, we have to make a decision about which 9 out of 18 vectors do we have to choose from and also how to arrange them before mixing. All these variations result in different outputs.

### B. Pre-trained feature directions

The basic arithmetic method, as discussed in A does not guarantee that the resultant vector will only contain the desired feature but only estimates it. To solve this problem, we use a new method. In this method, we aim to find planes that separate the features from on another in the latent space. After finding and selecting a plane between features, we can find a direction that is normal to the plane and then can start moving the vector in that direction. For example, if we consider 2 features i.e., glasses and no glasses. We have to find a plane that separates these 2 features. Then if we find a direction normal to the plane, we can move the vector toward either direction i.e., glasses and no glasses. This results in either the feature glasses begin removed or added according to the direction of the movement. This plane can be found by creating a dataset for a binary feature and then training a classification model on it such as a logistic regression or SVM. The learned function of the model provides the data needed for the plane

### C. Encoding real images

In all the experiments above, we can see that the images generated would be either from the dataset or be completely new. The biggest applications of GAN's are more visible if we can somehow generate an image that belongs to a real person. As we now know that the latent vector decides what type of image will be generated, our aim should be to find the latent vector that would, when passed through the generator, generate a target image. There are 2 ways to achieve this goal. The most obvious is to have an encoder network, that takes as input the image, and by finding the underlying features maps the image to a vector. The second is to perform gradient descent on the latent vector itself by using a loss between the current generated image and the target image.

The encoder network architecture is similar to the inverse of the generator of the GAN. It is trained by using a dataset of latent vectors and their images. The biggest problem with this method is that it takes a lot of data to train and in some cases the network might nor generalize to data outside its dataset. Another problem with this model is that the images have to be perfectly aligned before encoding to get best results. In some cases, alignment is not necessary but the output improves if the image has a pose similar to that of the dataset.

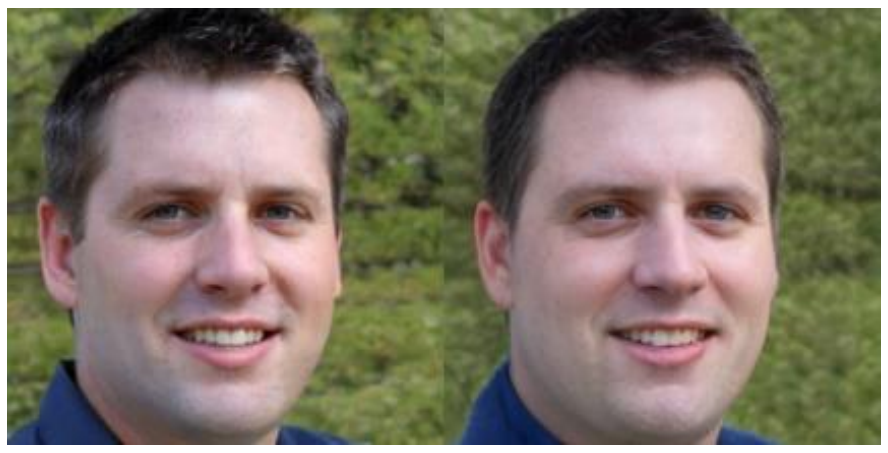

Figure 8: The above figure shows the result of the encoder technique to find a latent vector that generates the original image. As we can see the result from this technique is not that accurate. There are major visible differences between the images but this method takes very little time to execute.

The next method is to perform gradient descent on the latent vector. In this method, an optimization algorithm such as gradient descent is used to optimize the latent vector values for a loss defined between the target image and the image generated by the GAN by using the current latent vector. By doing this the weights of the model are not affected but only the latent vector is optimized and starts to slowly move in the correct direction. The loss is defined by using some of the layers of a pretrained classifier network. These layers contain the high-level features of the input image and hence provide better accuracy as compared to a simple l2 pixel wise loss.

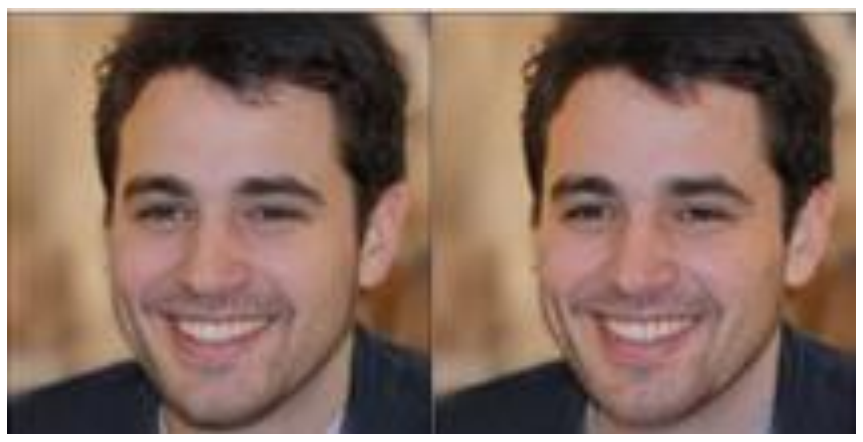

Figure 9: The above figure shows the result of the optimization technique to find a latent vector that generates the original image. This process gives very good results but needs a lot of time. As we can see in the above image, both the images look very similar but do have some small differences.

## IV. Results and Disccusions

After our experiments, we conducted a short survey among our peers to compare our results with the results of models that perform a similar task. In this survey, the participants were shown the images and asked to assign a score out of 10, on 4 parameters – Image quality, Original Feature Retention, New Feature Addition, No. of Editable Features. The first parameter was used to score the model on the quality of generated images. The 2nd parameter was used to score the model based on how well the model retained/remembered the features from the original image. The 3rd parameter was used to score the model on how well it added new features to the original image. The last parameter was used to score the model on the no of features it was capable of changing/adding. The resultant scores can be seen in Figure 10.

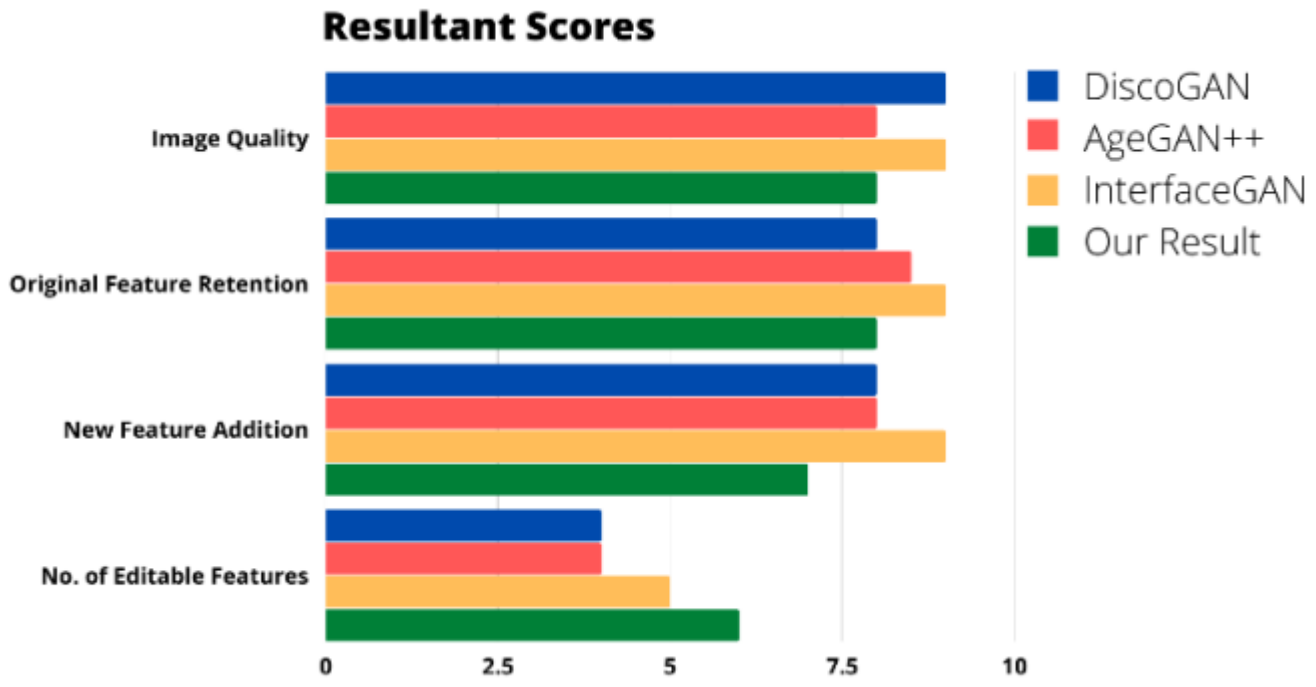

Figure 10: This figure shows that the interface GAN network works the best among these.

## V. Conclusion

In recent years, GANs have evolved from a novel idea to a new technique that has the potential to revolutionize what is possible with deep learning. The interest in generative adversarial networks is not only due to their ability to create data from a latent space into a data space, but also that they have the potential for making use of the vast quantities of unlabeled data. In this paper we have attempted to give an overview of this technique and its applications. We have shown that GANs can accomplish state-of-the-art performance and demonstrated that their inner workings can provide insight into future directions for research in both deep learning and representation learning. We have reviewed how GANs can be applied to a range of applications including photorealistic images, text generation, and even synthesis of human poses. Within the subtleties of GAN training, there are many opportunities for developments in theory and algorithms, and with the power of deep networks, there are vast opportunities for new applications.